\documentclass[letterpaper]{article} 
\usepackage{aaai24}  
\usepackage{times}  
\usepackage{helvet}  
\usepackage{courier}  
\usepackage[hyphens]{url}  
\usepackage{graphicx} 
\usepackage{natbib}  
\usepackage{caption} 
\usepackage{algorithm}
\usepackage{algorithmic}

\usepackage{newfloat}
\usepackage{listings}

\DeclareCaptionStyle{ruled}{labelfont=normalfont,labelsep=colon,strut=off} 
\floatstyle{ruled}
\newfloat{listing}{tb}{lst}{}
\floatname{listing}{Listing}
\usepackage{amsmath}
\usepackage{amssymb}
\usepackage{xspace}
\makeatletter
\DeclareRobustCommand\onedot{\futurelet\@let@token\@onedot}
\def\@onedot{\ifx\@let@token.\else.\null\fi\xspace}
\def\eg{\emph{e.g}\onedot}

\makeatother

\title{VIXEN: Visual Text Comparison Network for Image Difference Captioning}
\author {
    Alexander Black\textsuperscript{\rm 1},
    Jing Shi\textsuperscript{\rm 2},
    Yifei Fan\textsuperscript{\rm 2},
    Tu Bui\textsuperscript{\rm 1},
    John Collomosse\textsuperscript{\rm 1,2}
}
\affiliations {
    \textsuperscript{\rm 1}CVSSP, University of Surrey\\
    \textsuperscript{\rm 2}Adobe Research\\
    \{alex.black\textbar t.v.bui\}@surrey.ac.uk, \{jingshi\textbar yifan\textbar collomos\}@adobe.com
}

\usepackage{bibentry}

\begin{document}

\maketitle

\begin{figure*}[ht]
    \centering
    \includegraphics[width=0.95\linewidth]{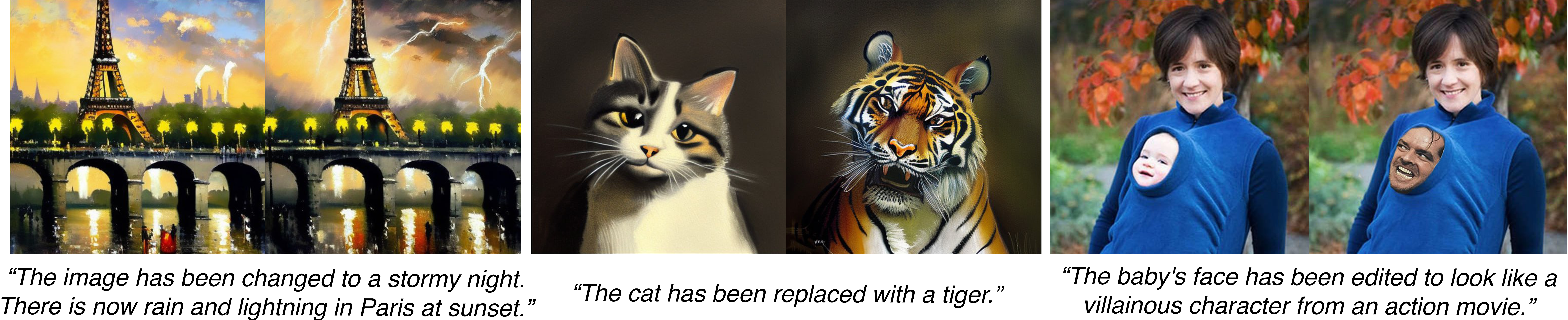}
    \caption{Visual change summarization produced by VIXEN for original-manipulated image pairs. VIXEN is able to observe both background (left) and main subject (mid) changes as well as generalize to other datasets (right).}
    \label{fig:teaser}
\end{figure*}

\begin{abstract}
We present VIXEN -- a technique that succinctly summarizes in text the visual differences between a pair of images in order to highlight any content manipulation present. Our proposed network linearly maps image features in a pairwise manner, constructing a soft prompt for a pretrained large language model. We address the challenge of low volume of training data and lack of manipulation variety in existing image difference captioning (IDC) datasets by training on synthetically manipulated images from the recent InstructPix2Pix dataset generated via prompt-to-prompt editing framework. We augment this dataset with change summaries produced via GPT-3.  We show that VIXEN produces state-of-the-art, comprehensible difference captions for diverse image contents and edit types, offering a potential mitigation against misinformation disseminated via manipulated image content. Code and data are available at \url{http://github.com/alexblck/vixen}
\end{abstract}

\section{Introduction}

Image manipulation often forms the basis for fake news and misinformation. This threat may be countered by tools that encourage users to reflect upon the provenance and content of images.  Given the reactionary nature of sharing, such tools should be intuitively comprehensible to enable users to make fast, informed trust decisions \cite{witness}.

This paper contributes VIXEN -- a method for intuitively summarizing the visual change between a pair of images using a short passage of text.  Emerging open standards (\eg C2PA \cite{c2pa}) describe provenance frameworks that match images circulating in the wild to a federated database of originals using perceptual hashing methods \cite{icn,oscarnet,vpn,pizzi}. VIXEN presents a comprehensible way to review any image manipulation evidenced by such a matching  (Fig.~\ref{fig:teaser}).  

Image difference captioning (IDC) is typically addressed by representations that seek to model the spatial-semantic distribution of concepts present in a scene -- for example, the relative positions of objects in CCTV footage \cite{spot-the-diff}, or of primitive geometric shapes \cite{clevr}. More complex kinds of  manipulations require expertise to construct and thus can not be easily scaled up in volume \cite{relatt}. To this end, we make three technical contributions:

\begin{enumerate}
\item{\textbf{Cross-modal image differencing.} We present a novel image differencing concept comprising a 2-branch GPT-J architecture to embed and compare facts derived from the image pair using CLIP-based image encoding.  The model generates text conditioned on that comparison to explain salient changes between the image pair. We show our textual explanations to be succinct and comprehensible to non-expert users and to be quantitatively closer to ground-truth edit captions than state-of-the-art captioning methods.}

\item{\textbf{Synthetic Edit Training.} We propose a synthetic pair-wise training framework for our VIXEN leveraging recent prompt2prompt (P2P) and language-based image editing (LBIE) approaches to supervise fine-tuning on generative image content, showing good generalization to unseen content.}

 \item{\textbf{Augmented IP2P Dataset.} We release an augmentation of the recent InstructPix2Pix (IP2P) dataset with synthetic change captions generated via GPT-3 as a basis for training and evaluating VIXEN.}
\end{enumerate}

We demonstrate that VIXEN achieves higher performance than prior image difference captioning methods and is able to generalize to multiple datasets.

\section{Related Work}
Image difference captioning (IDC) is closely related to image captioning and visual question answering, both requiring a visual understanding system to model images and a language understanding system capable of generating syntactically correct captions. The revolution of IDC in recent years depends heavily on the advent of visual and text modeling approaches, together with cross-domain learning techniques that bridge the representation gap between them.

Early visual content modeling approaches employ global CNN features such as VGG \cite{donahue2015long} and ResNet \cite{rennie2017self} as input signals to the text generation models, leveraging the semantically rich and compact representations deliverable from these models. To better capture multi-object representations and their relation, regional modeling methods are developed \cite{lu2017knowing,gu2018stack,anderson2018bottom,huang2019adaptively}. In some, images are gridded into non-overlapping patches upon which CNN features are extracted, others instead use outputs from an early layer of a pretrained ResNet model to effectively capture spatial features in a grid fashion. In contrast, \cite{cornia2020meshed,anderson2018bottom,huang2019adaptively} employ Region Proposal Network (RPN) to extract features from potential candidate objects, offering better alignment with semantic objects mentioned in the paired captions. Alternative approaches include graph-based \cite{yang2019auto} and tree-based networks \cite{yao2019hierarchy} to capture the relations of objects at multiple levels of granularity.  


For a long time RNN/LSTM \cite{graves2012long} have been used to model text due to its inherent sequential properties. Single-layer RNN \cite{vinyals2015show,mao2014deep} or double-layer LSTM \cite{donahue2015long,anderson2018bottom,yao2019hierarchy} are employed along with various techniques to integrate image features deeper into the recurrent process, including additive attention \cite{stefanini2022show}. During inference, captions are generated in an autoregressive fashion -- the prediction of a word is conditioned on all previous words. While this improves linguistic coherence, RNN/LSTM-based approaches struggle in modeling long captions. This problem is levitated in recent transformer-based approaches thanks to its full-attention mechanism \cite{luo2021dual,wang2021simvlm,cornia2020meshed}. More advanced transformer-based approaches such as BERT \cite{devlin2018bert}, GPT \cite{gpt-3} and LLaMA \cite{llama} have been successfully applied in various visual-language tasks \cite{hu2022scaling,mokady2021clipcap,gao2023llama,zhang2021vinvl,li2020oscar}.    


Visual language modeling aims to bridge the gap between image/video and text representations for specific tasks such as joint embedding (\eg CLIP \cite{CLIP} and LIMoE \cite{limoe} for cross-domain retrieval), text-to-image (\eg Stable Diffusion \cite{sd} for text-based image generation, InstructPix2Pix \cite{ip2p} for image editing) and image-to-text (\eg visual question answering \cite{flamingo,wang2021simvlm}, visual instructions \cite{gao2023llama,palme}). In the context of image captioning, image-text mapping strategies can be categorized into two research strands. The first strand involves the early fusion of image and text features for better alignment between image objects and words \cite{frozen,mokady2021clipcap,wang2021simvlm,li2020oscar}. These methods adopt BERT-like training strategies to input a pair of image and masked caption to the masked words. At inference, the input caption is simply replaced by a start token or a prefixed phrase \eg `A picture of'. The second research strand focuses on learning a direct transformation from image to text embedding. Early  CNN-based approaches feed image features as the hidden states of the LSTM text modules \cite{donahue2015long,vinyals2015show,yao2019hierarchy,karpathy2015deep,rennie2017self} while later transformer-based methods favor cross-attention \cite{luo2021dual,cornia2020meshed}. Recently in both research strands, there has been a trend of leveraging powerful pretrained large language and vision models to learn a simple mapping between two domains \cite{LIMBER,MAGMA,blip2,frozen,mokady2021clipcap}. 

Image difference captioning is a form of image captioning in which the caption would ideally ignore common objects between images and rather highlight subtle changes between them. As the first work addressing IDC, Spot-the-Diff \cite{spot-the-diff} identifies potential change clusters and models them using an LSTM-based network. Their work relies on the difference between two input images at the pixel level, therefore sensitive to noises and geometric transformations. DUDA \cite{DUDA} instead computes image difference at CNN semantic level, improving the robustness against slight global changes. In M-VAM \cite{M-VAM} and VACC \cite{kim2021agnostic}, a view-point encoder is proposed to mitigate potential view-point difference and VARD \cite{VARD} proposes a viewpoint invariant representation network to explicitly capture the change. Meanwhile, \cite{sun2022bidirectional} uses bidirectional encoding to improve change localization and NCT \cite{NCT} aggregates neighboring features with a transformer. These methods mostly focus on image modality and take advantage of benchmark-specific properties, such as near-identical views in Spot-the-Diff \cite{spot-the-diff} or synthetic scenes with limited objects and change types (color, texture, add, drop, remove) in CLEVR \cite{DUDA}. More recently, IDC-PCL \cite{IDC} and CLIP4IDC \cite{CLIP4IDC} adopt BERT-like training strategies to model difference captioning language, achieving state-of-art performance. 



\begin{figure*}[ht]
    \centering
    \includegraphics[width=0.9\linewidth]{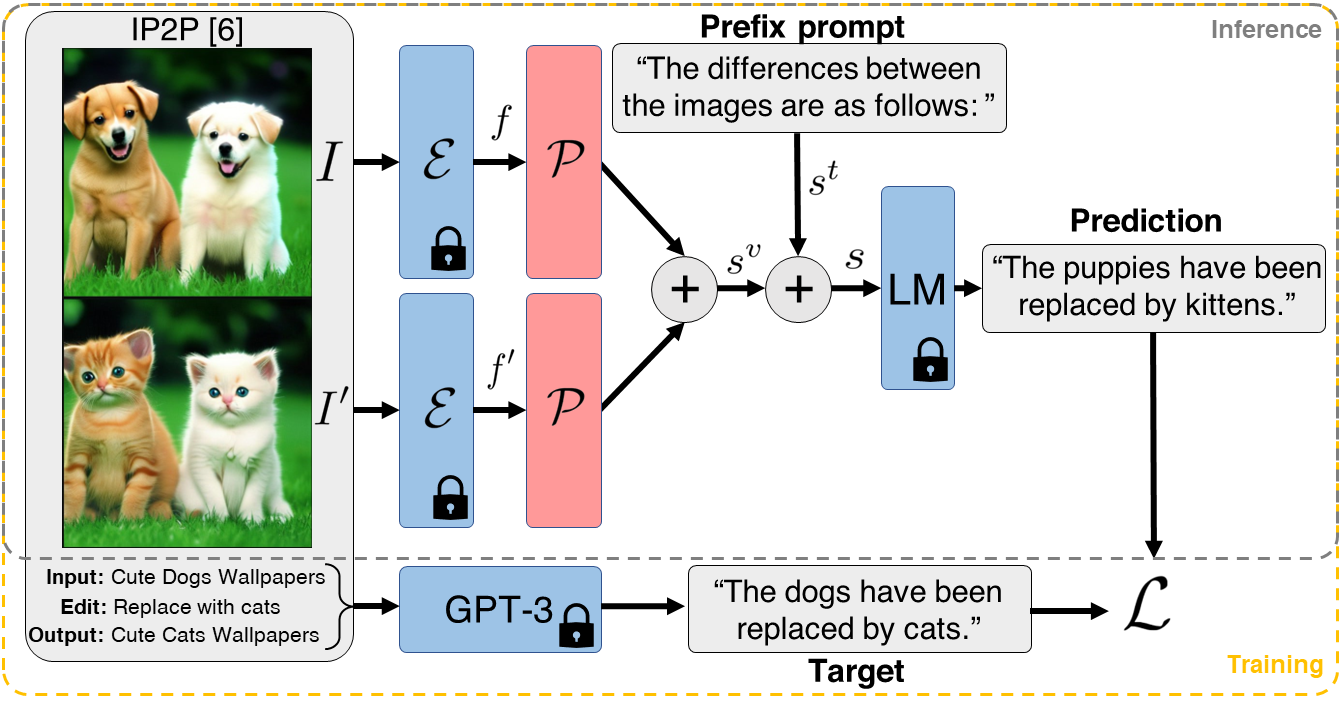}
    \caption{Model architecture and data captioning augmentation pipeline diagram. We use a pre-trained image encoder network $\mathcal{E}$ to produce a representation of two images. Both of these are projected into the input space of a large language model (LM) by a trained linear projection layer $\mathcal{P}$. Frozen layers are marked in blue, trainable in red.}
    \label{fig:arch}
\end{figure*}

\section{Methodology}
Our proposed method relies on synthetically generated image pairs and associated difference captions. We describe the creation process of visual and textual components of the dataset, details of the architecture of our proposed approach and training details necessary to reproduce the results. 

\subsection{Data Generation}

To train our proposed approach, we require a large dataset of image pairs, each annotated with a summary of the changes between them. We propose using images generated by stable diffusion \cite{sd} and edited with prompt-to-prompt \cite{p2p} using the pipeline presented in InstructPix2Pix \cite{ip2p} (IP2P). One of our contributions is the introduction of difference summary captions to IP2P images, generated using GPT-3 \cite{gpt-3} in a few-shot learning fashion. 

The InstructPix2Pix dataset is generated using the prompt-to-prompt editing framework, which provides text-based editing capabilities for synthesized images by injecting the attention maps associated with a specific word in the prompt to control the attention maps of the edited image. Therefore, all that is required to generate an image pair is two textual prompts with slight differences. IP2P uses a fine-tuned GPT-3 language model to generate plausible edits based on real input captions from LAION \cite{laion}. In addition to the image pairs and captions the dataset also contains an instruction that describes what edits have  to be  applied  in order to generate the output image.

While these instructions are sufficient for the original InstructPix2Pix task of text-based image editing, they often  omit the information regarding the input content. For example, for the prompt pair "a photo of a cat"/"a  photo of  a dog", the edit instruction might be "as a dog" or "turn it into a dog". We aim to summarize the changes by referencing both the original and edited image contents, therefore the desirable edit summarization caption would be "the cat has been replaced by a dog". To achieve this, we use GPT-3 language model in a few-shot learning fashion by including several examples of input-output-instruction-summary quadruplets where summary captions are constructed manually. While IP2P uses a fine-tuned GPT-3 to generate the instruction and second image captions, we have found the fine-tuning unnecessary in our case. Since our task does not require creativity from the model, but rather summarization of the input information, the pre-trained 'davinci' version of GPT-3 is enough to produce the captions needed.

\begin{figure*}[ht!]
    \centering
    \includegraphics[width=0.9\linewidth]{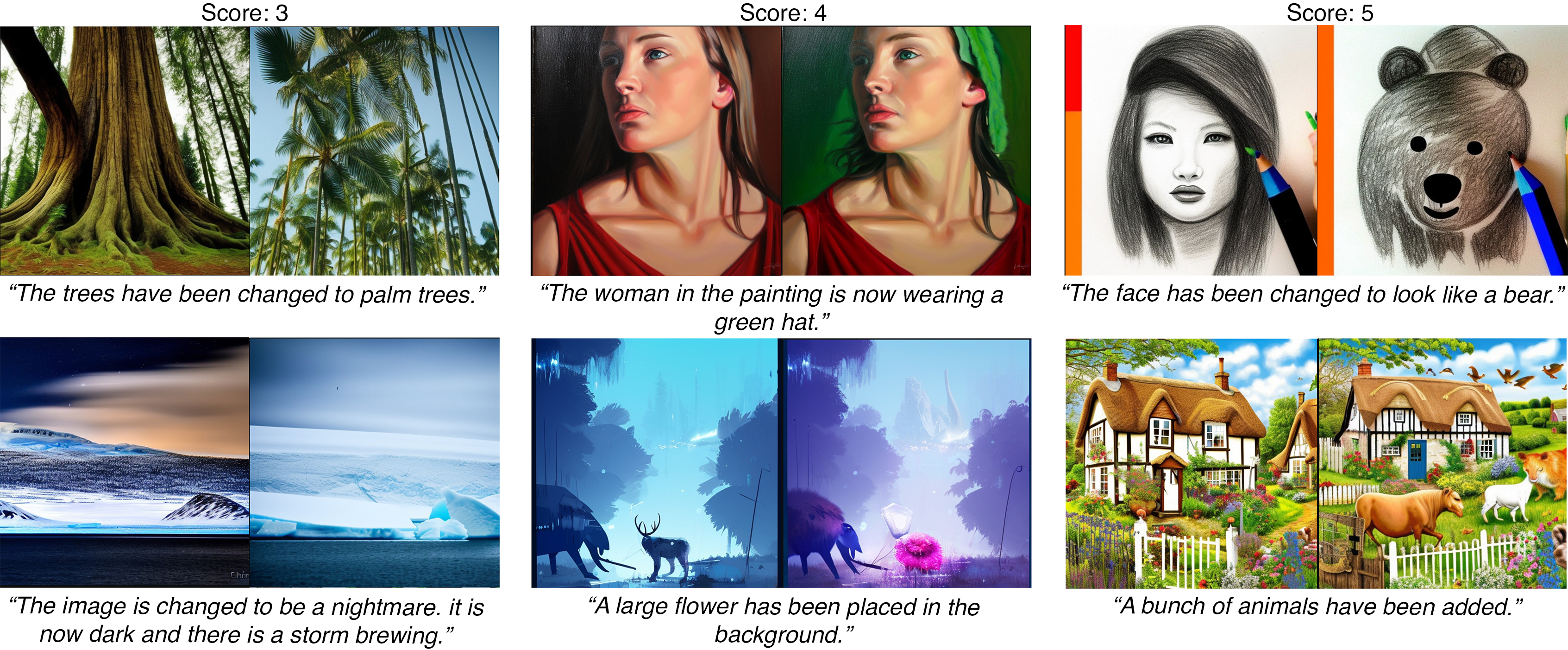}
    \caption{Image-caption pairs with an average correspondence score of 3  (left): may contain global changes when only local ones are  expected  (top) or fail to  produce desired edits due to vague captioning (bot); 4 (mid): partially satisfy the caption, occasionally only some properties are realized correctly (top) or an existing object is replaced rather than added to the background (bot); 5  (right): mostly faithful to the depicted edits.}
    \label{fig:mturk}
\end{figure*}

\subsection{Architecture}
\label{sec:method:arch}

Our image captioning approach is inspired by \cite{LIMBER}, which uses a trainable linear mapping between the image encoder and a large language model. However, instead of passing the projected embeddings of a single image to the language model, we project the embeddings of two images and concatenate them before feeding them into the language model. This  architecture is illustrated in Figure~\ref{fig:arch}. Given a source image $I$ and its edited version $I'$ we use an image encoder $\mathcal{E}$ to extract image feature maps 
\begin{equation}
    f=\mathcal{E}(I);\; f'=\mathcal{E}(I') \in \mathbb{R}^{k \times h},
\end{equation}
where $h$ is the size of feature maps and $k$ is the prompt sequence length.
We use a fully-connected layer $\mathcal{P}$ to linearly project the image features into dimensionality of a language model input $e$, creating a soft prompt $s^v$:
\begin{equation}\label{eq:soft}
    s^v = [\mathcal{P}(f), \mathcal{P}(f')]  \in \mathbb{R}^{2k \times e},
\end{equation}
where $[,]$ denotes concatenation.
Finally, we append a prefix $s_t$ made of embedding of tokens for \textit{"The differences between the images are as follows: "/"Edit instructions:"} to the visual prompt $s^v$ to obtain the final prompt $s = [s^v, s^t]$ used for generating the summarization text. 

We explore two options for $\mathcal{E}$. Firstly, following \cite{LIMBER, MAGMA}, we use CLIP RN50x16 as $\mathcal{E}$. The feature map before the pooling layer has dimensions $12\times12\times3072$, flattened to $k\times h=144 \times 3072$. Secondly, we use ViT-g, followed by a Q-Former from BLIP-2 \cite{blip2}. In this case sequence length $k=257$. We refer to CLIP and Q-Former versions of VIXEN as \textbf{VIXEN-C} and \textbf{VIXEN-Q}, respectively. For the language model, we use GPT-J\cite{gpt-j}, which has input space dimensionality $l=4096$. Consequently, for both configurations of $\mathcal{E}$, our linear projection layer $\mathcal{P}$ has input and output dimensions $h=3072$ and $l=4096$, respectively. The loss for the captioning task objective is defined as
\begin{equation}
    \mathcal{L} = -\sum^{m}_{i=1}l(s^v, s^{t}_1, \ldots, s^{t}_i),
\end{equation}
where $m$ is a variable token length and $l$ is next-token log-probability conditioned on the previous sequence elements

\begin{equation}
    l(s^v, s^{t}_1, \ldots, s^{t}_i) = \log p(t_i| x, t_1, \ldots, t_{i-1}).
\end{equation}




\subsection{Training}
\label{sec:method:training}
During training, we may provide distractor image pairs with no changes present by providing the same image as both inputs $I = I'$. The frequency of the presence of distractor images is determined by probability $p_d$. In such cases, the target difference summary text is chosen at random from a list of pre-defined sentences, all synonymous with "there is no difference". For all our models we first train with $p_d=0$ for two epochs, followed by two more epochs with $p_d=0.5$. Total training time is approximately  100 hours on a single A100 GPU. We use gradient accumulation to train with an effective batch size of 2048 and optimize the loss using AdamW optimizer with $\beta_1 = 0.9, \beta_2 = 0.98$ and  weight decay 0.05. For baseline approaches CLIP4IDC and IDC, we implemented dataloaders for our dataset, precomputed all necessary supporting data (e.g., ResNet-101 features, negative samples, and a vocabulary dictionary for IDC) and followed their standard two-step training pipeline with default hyperparameters specified in the GitHub repos. 

\section{Experiments}

\begin{table*}[ht]
\centering
\resizebox{\textwidth}{!}{%
\begin{tabular}{llllllllllllllll}
\hline
\multicolumn{1}{l|}{\textbf{Method}} & \multicolumn{3}{c|}{\textbf{MPNet}} & \multicolumn{3}{c|}{\textbf{B@4}} & \multicolumn{3}{c|}{\textbf{C}} & \multicolumn{3}{c|}{\textbf{M}} & \multicolumn{3}{c}{\textbf{R}} \\ \hline
\multicolumn{16}{c}{Instruct Pix2Pix} \\ \hline
\multicolumn{1}{l|}{} & @3 & @4 & \multicolumn{1}{l|}{@5} & @3 & @4 & \multicolumn{1}{l|}{@5} & @3 & @4 & \multicolumn{1}{l|}{@5} & @3 & @4 & \multicolumn{1}{l|}{@5} & @3 & @4 & @5 \\ \hline
\multicolumn{1}{l|}{VIXEN-Q (ours)} & \underline{56.9} & \underline{59.1} & \multicolumn{1}{l|}{\textbf{62.3}} & \underline{16.5} & \textbf{18.5} & \multicolumn{1}{l|}{\textbf{20.8}} & \underline{80.3} & \underline{93.9} & \multicolumn{1}{l|}{\textbf{134.9}} & 17.1 & 18.4 & \multicolumn{1}{l|}{\underline{20.6}} & \underline{38.0} & \underline{40.1} & \textbf{42.2} \\
\multicolumn{1}{l|}{VIXEN-C (ours)} & \textbf{59.3} & \textbf{61.4} & \multicolumn{1}{l|}{\underline{61.5}} & \textbf{16.8} & \underline{18.2} & \multicolumn{1}{l|}{\underline{19.2}} & \textbf{96.6} & \textbf{107.0} & \multicolumn{1}{l|}{\underline{126.1}} & \underline{17.6} & \underline{18.6} & \multicolumn{1}{l|}{19.6} & \textbf{39.2} & \textbf{40.8} & \underline{39.3} \\
\multicolumn{1}{l|}{CLIP4IDC} & 56.8 & 58.3 & \multicolumn{1}{l|}{60.7} & 15.8 & 17.3 & \multicolumn{1}{l|}{17.7} & 58.8 & 71.0 & \multicolumn{1}{l|}{114.7} & \textbf{20.9} & \textbf{22.5} & \multicolumn{1}{l|}{\textbf{23.3}} & 33.3 & 35.1 & 34.0 \\
\multicolumn{1}{l|}{IDC} & 38.3 & 38.6 & \multicolumn{1}{l|}{37.4} & 8.2 & 8.8 & \multicolumn{1}{l|}{7.7} & 4.4 & 5.0 & \multicolumn{1}{l|}{5.6} & 16.0 & 16.8 & \multicolumn{1}{l|}{16.5} & 29.1 & 30.0 & 27.7 \\ \hline
\multicolumn{16}{c}{PSBattles} \\ \hline
\multicolumn{1}{l|}{VIXEN-Q (ours)} & \multicolumn{3}{c|}{\textbf{45.1}} & \multicolumn{3}{c|}{\textbf{5.8}} & \multicolumn{3}{c|}{\underline{7.5}} & \multicolumn{3}{c|}{\textbf{11.0}} & \multicolumn{3}{c}{\textbf{22.2}} \\
\multicolumn{1}{l|}{VIXEN-C (ours)} & \multicolumn{3}{c|}{\underline{40.3}} & \multicolumn{3}{c|}{\underline{4.5}} & \multicolumn{3}{c|}{\textbf{7.7}} & \multicolumn{3}{c|}{9.5} & \multicolumn{3}{c}{20.5} \\

\multicolumn{1}{l|}{CLIP4IDC} & \multicolumn{3}{c|}{32.7} & \multicolumn{3}{c|}{3.2} & \multicolumn{3}{c|}{5.0} & \multicolumn{3}{c|}{\underline{10.1}} & \multicolumn{3}{c}{\underline{21.7}} \\
\multicolumn{1}{l|}{IDC} & \multicolumn{3}{c|}{27.0} & \multicolumn{3}{c|}{1.0} & \multicolumn{3}{c|}{0.7} & \multicolumn{3}{c|}{9.2} & \multicolumn{3}{c}{19.5} \\ \hline
\multicolumn{16}{c}{Image Editing Request} \\ \hline
\multicolumn{1}{l|}{VIXEN-Q (ours, FT)} & \multicolumn{3}{c|}{\underline{50.1}} & \multicolumn{3}{c|}{7.9} & \multicolumn{3}{c|}{35.4} & \multicolumn{3}{c|}{14.4}              & \multicolumn{3}{c}{33.5}\\
\multicolumn{1}{l|}{VIXEN-C (ours, FT)} & \multicolumn{3}{c|}{\bf 52.5} & \multicolumn{3}{c|}{\underline{8.6}}      & \multicolumn{3}{c|}{\bf38.1}          & \multicolumn{3}{c|}{\bf15.4}           & \multicolumn{3}{c}{\bf42.5}\\
\multicolumn{1}{l|}{VARD       } & \multicolumn{3}{c|}{-}   & \multicolumn{3}{c|}{\bf10.0} & \multicolumn{3}{c|}{\underline{ 35.7}} & \multicolumn{3}{c|}{14.8}  & \multicolumn{3}{c}{39.0}\\
\multicolumn{1}{l|}{CLIP4IDC    }& \multicolumn{3}{c|}{-} & \multicolumn{3}{c|}{8.2}           & \multicolumn{3}{c|}{32.2}             & \multicolumn{3}{c|}{14.6}  & \multicolumn{3}{c}{\underline{40.4}}\\
\multicolumn{1}{l|}{NCT         \ }& \multicolumn{3}{c|}{-}     & \multicolumn{3}{c|}{8.1}                 & \multicolumn{3}{c|}{34.2}             & \multicolumn{3}{c|}{\underline{15.0}}              & \multicolumn{3}{c}{38.8}\\
\multicolumn{1}{l|}{BiDiff      }& \multicolumn{3}{c|}{-}   & \multicolumn{3}{c|}{6.9}                     & \multicolumn{3}{c|}{27.7}             & \multicolumn{3}{c|}{14.6}              & \multicolumn{3}{c}{38.5}\\
\multicolumn{1}{l|}{DUDA       }& \multicolumn{3}{c|}{-}     & \multicolumn{3}{c|}{6.5}                    & \multicolumn{3}{c|}{27.8}             & \multicolumn{3}{c|}{12.4}              & \multicolumn{3}{c}{37.3}\\
\multicolumn{1}{l|}{rel-att  }& \multicolumn{3}{c|}{-}      & \multicolumn{3}{c|}{6.7}                     & \multicolumn{3}{c|}{26.4}             & \multicolumn{3}{c|}{12.8}              & \multicolumn{3}{c}{37.4}\\\hline
\end{tabular}
}
\caption{Image difference captioning performance on IP2P,  PSBattles and Image Editing Request datasets. Evaluated on semantic similarity (MPNet), BLEU-4 (B@4), CIDEr (C), METEOR (M) and ROUGE-L (R). For IP2P, performance is reported for  subsets at image-caption correspondence thresholds of 3,  4, 5.}\label{tab:eval_metrics}
\end{table*}

\begin{table}[ht]
\centering
\begin{tabular}{llllll}
\hline
\multicolumn{1}{l}{\textbf{Method}} & \multicolumn{1}{l}{\textbf{MPNet}} & \multicolumn{1}{l}{\textbf{B@4}} & \multicolumn{1}{l}{\textbf{C}} & \multicolumn{1}{l}{\textbf{M}} & \textbf{R} \\ \hline
\multicolumn{6}{c}{Instruct Pix2Pix} \\ \hline
\multicolumn{1}{l}{VIXEN-C (ours)} & \multicolumn{1}{l}{\textbf{59.3}} & \multicolumn{1}{l}{\textbf{16.8}} & \multicolumn{1}{l}{\textbf{96.6}} & \multicolumn{1}{l}{\textbf{17.6}} & \textbf{39.2} \\
\multicolumn{1}{l}{VIXEN-C p=0} & \multicolumn{1}{l}{54.4} & \multicolumn{1}{l}{15.4} & \multicolumn{1}{l}{88.5} & \multicolumn{1}{l}{16.1} & 35.9 \\ \hline
\multicolumn{6}{c}{PSBattles} \\ \hline
\multicolumn{1}{l}{VIXEN-C (ours)} & \multicolumn{1}{l}{\textbf{40.3}} & \multicolumn{1}{l}{\textbf{4.5}} & \multicolumn{1}{l}{\textbf{7.7}} & \multicolumn{1}{l}{\textbf{9.5}} & \textbf{20.5} \\
\multicolumn{1}{l}{VIXEN-C p=0} & \multicolumn{1}{l}{37.8} & \multicolumn{1}{l}{4.2} & \multicolumn{1}{l}{7.2} & \multicolumn{1}{l}{8.9} & 19.2 \\ \hline
\end{tabular}%
\caption{Impact of distractor images on performance of the model evaluated on semantic similarity (MPNet), BLEU-4 (B@4), CIDEr (C), METEOR (M) and ROUGE-L (R).}\label{tab:ablation}
\end{table}

\subsection{Data}
\label{sec:experiments:data}
We perform our main evaluation on a subset of the InstructPix2Pix \cite{ip2p} dataset, unseen by models during training. To ensure a high quality of the synthetically generated image-caption pairs, we score their correspondence via a user study. Additionally, we crowd-source annotations for a subset of images from the PSBattles \cite{psBattles} dataset and fine-tune and evaluate on Image Editing Request \cite{relatt}.

InstructPix2Pix dataset presents challenges due to its synthetically generated nature, as some of the edit summarization captions fail to accurately describe the changes made to the image pairs. This is mainly due to prompt-to-prompt occasionally generating images that do not depict the desired change accurately enough. This is further discussed in  the limitations section below and illustrated in Figure \ref{fig:limitations} (mid).  To ensure a reliable evaluation, we conducted a user study using Amazon Mechanical Turk (MTurk) on a sample of 5,000 images from the dataset. This results in a 837,466/93,052/5,000 train/validation/test splits. The study involved three participants per image-caption pair (95 unique participants) and aimed to rate the degree of correspondence between the image pair and its associated caption, using a scoring system from 1 (low) to 5 (high). The distribution of scores is 1: $5\%$, 2: $13\%$, 3: $26\%$, 4: $33\%$, 5: $24\%$ . Figure \ref{fig:mturk} shows random samples of the image-caption pairs for different score threshold values.

PSBattles is a dataset of images edited in Adobe Photoshop\textsuperscript{TM}, collected from the `Photoshopbattles’ subreddit. The dataset contains 10k original images, paired with several manipulated variants. There are 102k variants in total contributed by 31k artists. We randomly sample 100 image pairs for crowd-sourced annotation on MTurk and collect captions from 3 participants per image pair.

Image Editing Request is a dataset of realistic photographs, paintings and illustrations paired with instructions written by humans. It contains 4k images-annotations pairs and incorporates a wide variety of edits, including affine edits and crops that are not present in the other datasets.



\subsection{Metrics}
\label{sec:experiments:metrics}
We evaluate the performance of difference captioning methods using both traditional N-gram-based metrics (BLEU-4 \cite{bleu}, CIDEr \cite{cider}, METEOR \cite{meteor} and ROUGE-L \cite{rouge}), as well as semantic similarity metric based on a language transformer model. We have found that due to a larger diversity of images  and edits, the generated captions need to encompass a significantly larger vocabulary to accurately describe the changes. As a result, there are instances where the captions may not align word for word with the actual image differences, but they still convey a similar meaning. To account for this, we use a  semantic textual similarity metric. We define semantic textual similarity $S_\mathrm{sim}$ between the target $c$ and generated $c'$ summarizations
\begin{equation}
    S_\mathrm{sim} = \cos(E(c), E(c')),
\end{equation}
where $\cos(,) = \frac{\mathbf{A} \cdot \mathbf{B}}{\|\mathbf{A}\| \|\mathbf{B}\|}$ denotes cosine similarity and $E$ is a sentence transformer. We use MPNet \cite{MPNet} as the best-performing sentence transformer to map sentences to 768-dimensional normalized embeddings. 

We also  assess the quality of captions via a crowd-sourced study on Amazon Mechanical Turk (MTurk). Participants are presented with both the original and edited images. For each image pair, participants are tasked to choose one of the 4 captions, arranged in a random order. In case all four captions do not summarize the differences well enough, participants may choose  the 'none of the above' option. Each task is performed by 3 unique participants. The preference is considered to be given to a particular method if two or more participants have voted for it.

\begin{figure*}[ht!]
    \centering
    \includegraphics[width=0.9\linewidth]{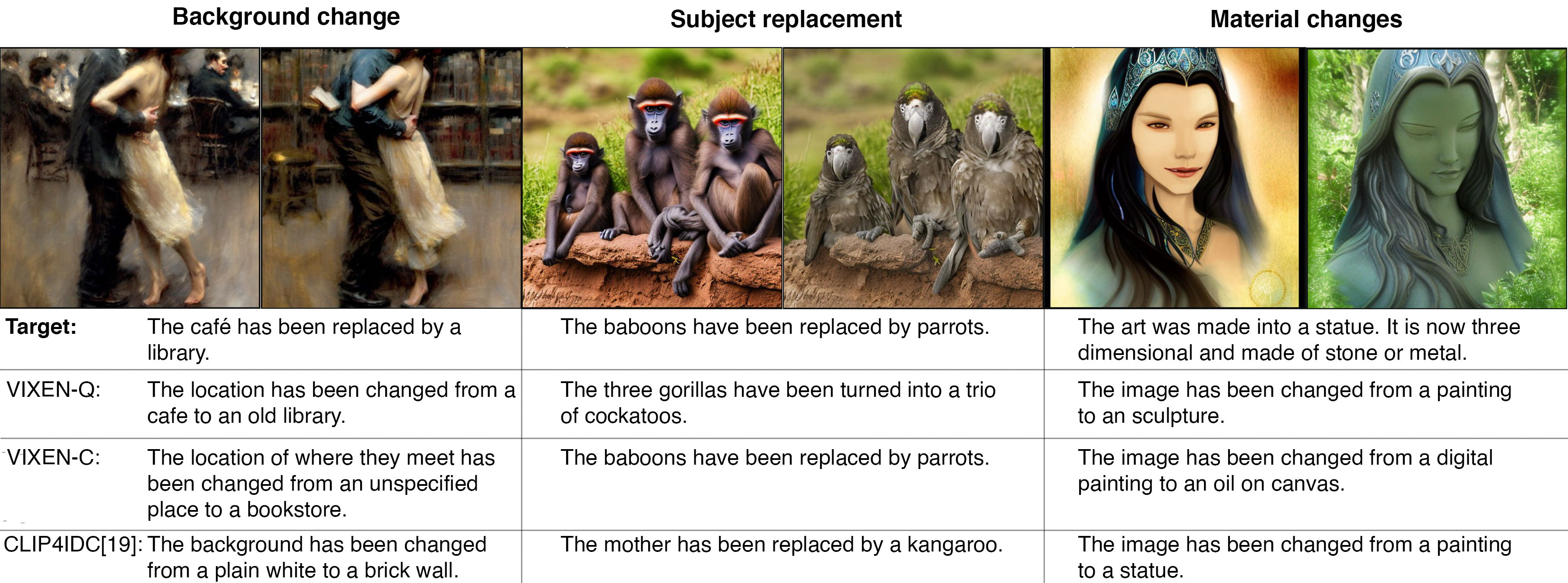} \\
    (a) InstructPix2Pix\\
    \includegraphics[width=0.9\linewidth]{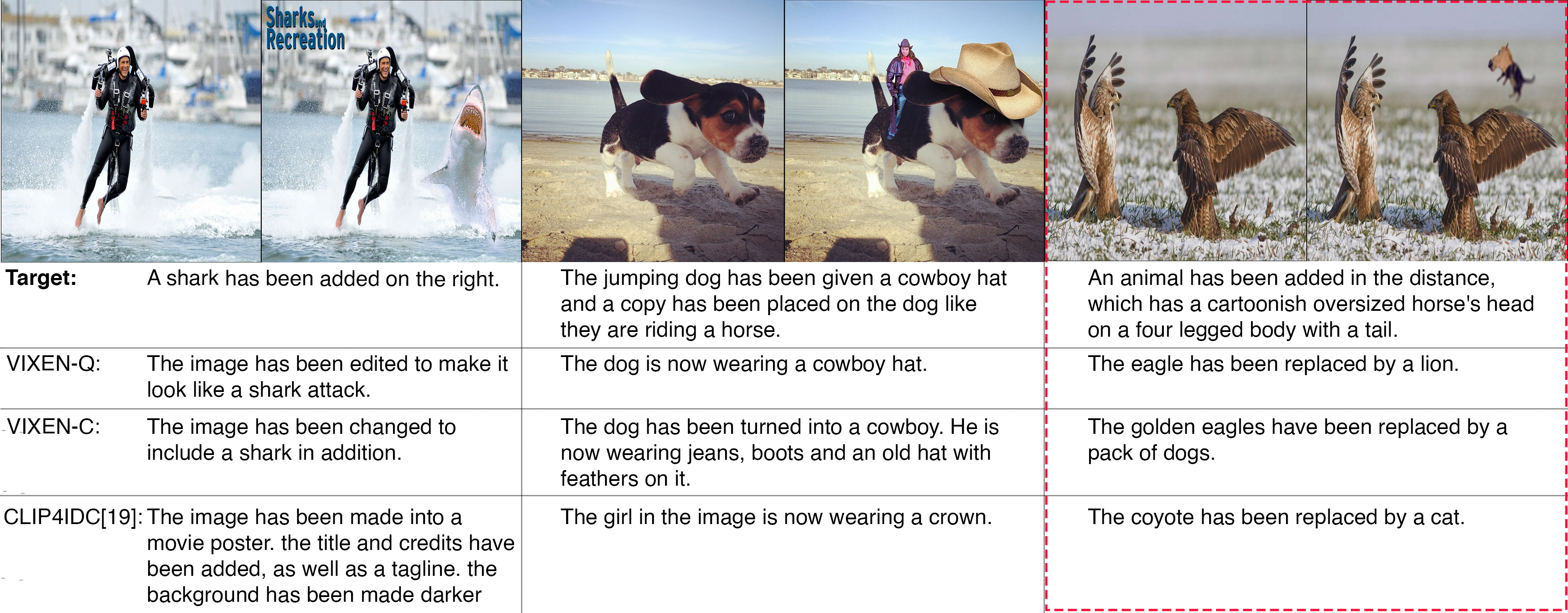} \\
    (b) PSBattles
    \caption{Examples of edit summarizations for global changes, object replacement and material changes produced by VIXEN and CLIP4IDC on  InstructPix2Pix (a) and PSBattles (b) datasets. Failure case marked with a dashed red box.}
    \label{fig:examples}
\end{figure*}

\subsection{Results}
For the proposed datasets, we compare the performance of VIXEN against two baselines, IDC \cite{IDC} and CLIP4IDC \cite{CLIP4IDC}. We train both of them on our augmented IP2P dataset, following the author's guidelines. For the IER dataset, we fine-tune on IER training set and compare against reported numbers of multiple baselines.

We report the evaluation results of evaluating both the proposed method as well as baselines in Table~\ref{tab:eval_metrics}, with examples shown in Figure \ref{fig:examples}. Our methods achieve a higher score in all metrics, except METEOR (IP2P), where CLIP4IDC scores higher than both proposed architectures. This indicates that VIXEN is more tuned towards precision, rather than recall of n-grams as METEOR heavily favors recall. For IP2P, results  are  reported at three different  correspondence thresholds.  For lower threshold values, the best results are obtained by VIXEN-C. VIXEN-Q seems to benefit the most  from threshold increase and outperforms other methods on pairs with a correspondence score  of 5. 

While all methods suffer significant performance drops when evaluated on a dataset from a different domain, VIXEN-Q shows a better ability to generalize to new data by scoring the highest on the PSBattles dataset. After fine-tuning the model on Image Editing Request, VIXEN-C outperforms previous methods on most metrics, except B@4 of VARD\cite{VARD}.


    
\begin{figure}[ht!]
    \centering
    \includegraphics[width=0.9\linewidth]{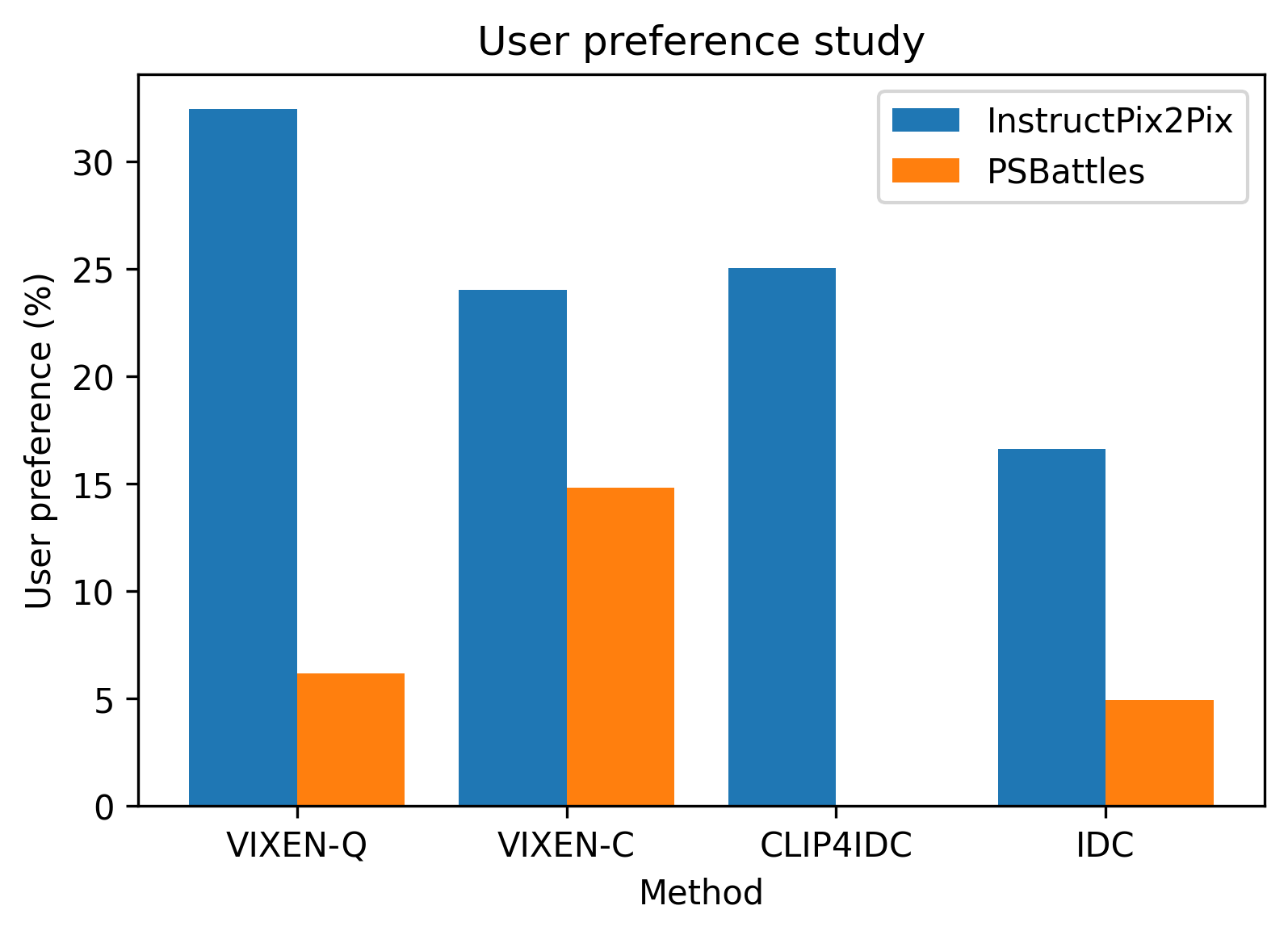}
    \caption{User preference study results. Study participants are shown an image pair and captions generated by four methods on IP2P and PSBattles datasets.}
    \label{fig:user_pref}
\end{figure}

The results of the crowd-sourced user preference study, shown in Figure \ref{fig:user_pref}, demonstrate that the users prefer difference captions generated by VIXEN more often than others. For the IP2P dataset, captions generated by VIXEN-Q and VIXEN-C obtained a majority vote in 32\% and 26\% of the cases, respectively, followed by CLIP4IDC and IDC with 24\% and 15\%. For the PSBattles dataset, the highest preference score is achieved by VIXEN-C with 15\% of the votes. Participants chose the 'None of the above' option in 75\% of  the cases for PSBattles, as opposed to just 2\% in IP2P. This indicates that generalization to new  data domains remains a challenging task.

\begin{figure*}[ht!]
    \centering
    \includegraphics[width=0.9\linewidth]{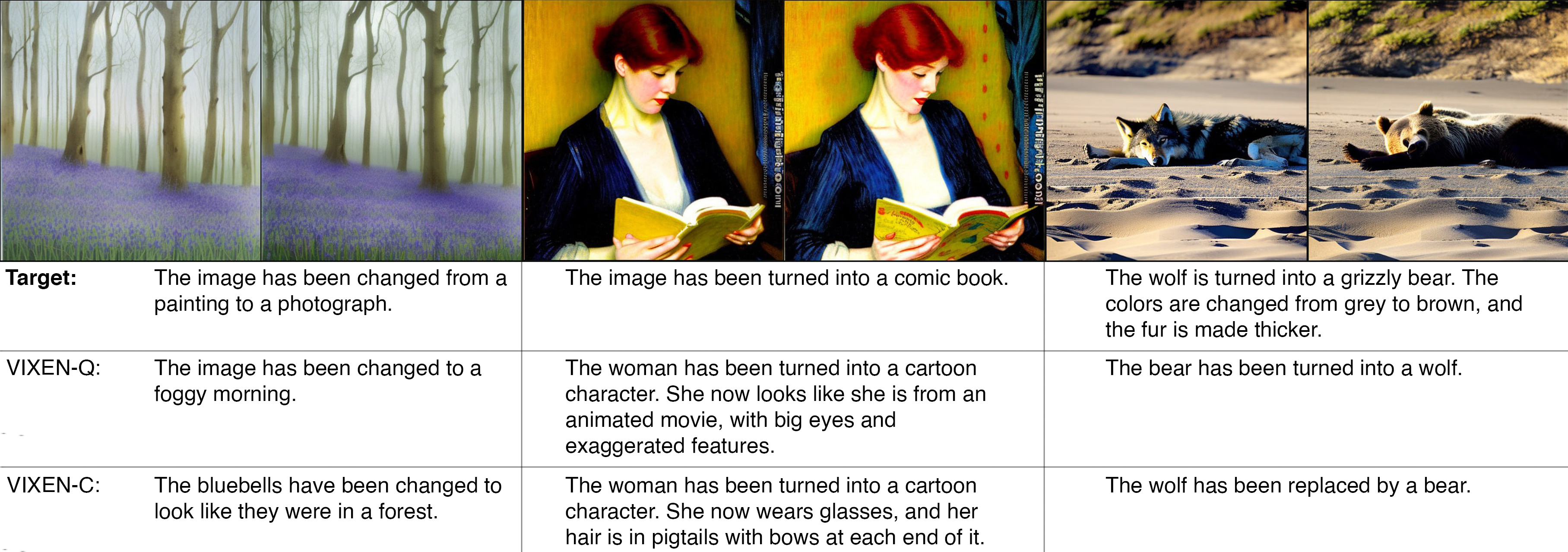}
    \caption{Limitations of the proposed method. Left: image captioning instead of difference captioning in case of unidentified edit. Middle: mismatch between target text-image pair and LM runoff. Right: edit described in reverse order.}
    \label{fig:limitations}
\end{figure*}

\begin{table}[ht!]
\centering
\begin{tabular}{lllll}
\hline
\multicolumn{1}{l}{\textbf{Fusion method}} & \multicolumn{1}{l}{\textbf{B@4}} & \multicolumn{1}{l}{\textbf{C}} & \multicolumn{1}{l}{\textbf{M}} & \textbf{R} \\ \hline
\multicolumn{5}{c}{Instruct Pix2Pix} \\ \hline
\multicolumn{1}{l}{Concatenation} & {\textbf{16.8}} & \multicolumn{1}{l}{\textbf{96.6}} & \multicolumn{1}{l}{\textbf{17.6}} & \textbf{39.2} \\
\multicolumn{1}{l}{Subtraction}  & \multicolumn{1}{l}{16.4} & \multicolumn{1}{l}{93.7} & \multicolumn{1}{l}{16.9} & 36.8 \\
\multicolumn{1}{l}{Addition}  & \multicolumn{1}{l}{16.2} & \multicolumn{1}{l}{90.8} & \multicolumn{1}{l}{17.3} & 37.5 \\ 
\multicolumn{1}{l}{Multiplication}  & \multicolumn{1}{l}{14.4} & \multicolumn{1}{l}{82.7} & \multicolumn{1}{l}{15.3} & 35.9 \\ 
\multicolumn{1}{l}{Mean}  & \multicolumn{1}{l}{10.7} & \multicolumn{1}{l}{63.9} & \multicolumn{1}{l}{12.4} & 33.6 \\ 
\hline
\end{tabular}%
\caption{Image feature fusion ablation of VIXEN-C}\label{tab:ablation2}

\end{table}

During inference we assume an input where one image is an edited  version of the other, but we demonstrate the benefits of having distractor same image pairs  during training. The possibility of no edits case makes it harder for the model to guess the right answer by memorizing the most frequent edits within the dataset. Table~\ref{tab:ablation} shows that setting the probability $p=0$ of the same image pairs during training of VIXEN-C yields worse results on both IP2P and PSBattles datasets.

Table \ref{tab:ablation2} shows performance results for different feature fusion strategies that redefine $s^v$ in Eq \ref{eq:soft}. We have observed that concatenation leads to slightly better performance than subtraction, addition or multiplication and taking the mean of two features causes a significant performance drop. This shows that retaining the information of both image features without degradation is important for the task.

\subsection{Limitations}
\label{sec:limitations}
In Figure~\ref{fig:limitations} we show examples of VIXEN's failure cases. We identify and discuss three main challenges. \textbf{Left} shows an example of a very  minor difference between the two images. In such cases, VIXEN occasionally resorts to captioning the image content instead of summarizing the differences. \textbf{Mid} shows a mismatch between the summary and generated images: an image pair with a slightly changed book cover, but the  target caption assumes that the style of the whole image has been changed to that of a comic book. As with other LLMs, VIXEN exhibits LM runoff: having identified a concept ("cartoon  character"), it might continue generating a text with a strong linguistic prior ("big eyes and exaggerated features"), absent in the images. \textbf{Right} shows that occasionally VIXEN may describe the differences between the images in a reversed order.


\section{Conclusion}
We presented VIXEN -- an image difference captioning approach that  provides textual descriptions of the manipulations applied to an image. We have augmented the InstructPix2Pix dataset of generated images with difference summarization captions generated by GPT-3 in order to train and evaluate VIXEN. We  have shown that VIXEN achieves higher performance than other image difference captioning methods. We have also demonstrated that, while VIXEN shows better generalizability to other  datasets, there is still a performance gap when switching from synthetic to real data. Future works might alleviate this by including a varied spread of manipulations types into the training set, including insertion, deletion and text edits, which the current generative pipelines struggle with.



\clearpage
\bibliography{aaai24.bib}

\end{document}